\documentclass[lettersize,journal]{IEEEtran}
\usepackage{amsmath,amsfonts}
\usepackage{algorithmic}
\usepackage{algorithm}
\usepackage{array}
\usepackage[caption=false,font=normalsize,labelfont=sf,textfont=sf]{subfig}
\usepackage{textcomp}
\usepackage{stfloats}
\usepackage{url}
\usepackage{verbatim}
\usepackage{graphicx}
\usepackage{cite}
\usepackage{multirow}

\begin{document}

\title{Emotion-Enhanced Multi-Task Learning with LLMs for Aspect Category Sentiment Analysis}


\author{Yaping Chai, Haoran Xie, Joe S. Qin
\thanks{This work was fully supported by a grant from the Research Grants Council of the Hong Kong Special Administrative Region, China (R1015-23); the Research Impact Fund by the Research Grants Council of Hong Kong (Project No. 130272); and the Faculty Research Grants (SDS24A8 and SDS24A19) and the Direct Grant (DR25E8) of Lingnan University, Hong Kong.
\emph{(Corresponding author: Haoran Xie.)}}

\thanks{Yaping Chai, Haoran Xie, and Joe S. Qin are with the School of Data Science, Lingnan University, Hong Kong (e-mail: yapingchai@ln.hk; hrxie@ln.edu.hk; joeqin@ln.edu.hk).}
}




\maketitle

\begin{abstract}
Aspect category sentiment analysis (ACSA) has achieved remarkable progress with large language models (LLMs), yet existing approaches primarily emphasize sentiment polarity while overlooking the underlying emotional dimensions that shape sentiment expressions. This limitation hinders the model’s ability to capture fine-grained affective signals toward specific aspect categories. To address this limitation, we introduce a novel emotion-enhanced multi-task ACSA framework that jointly learns sentiment polarity and category-specific emotions grounded in Ekman’s six basic emotions. Leveraging the generative capabilities of LLMs, our approach enables the model to produce emotional descriptions for each aspect category, thereby enriching sentiment representations with affective expressions. Furthermore, to ensure the accuracy and consistency of the generated emotions, we introduce an emotion refinement mechanism based on the Valence-Arousal-Dominance (VAD) dimensional framework. Specifically, emotions predicted by the LLM are projected onto a VAD space, and those inconsistent with their corresponding VAD coordinates are re-annotated using a structured LLM-based refinement strategy. Experimental results demonstrate that our approach significantly outperforms strong baselines on all benchmark datasets. This underlines the effectiveness of integrating affective dimensions into ACSA.
\end{abstract}


\section{Introduction}

Aspect category sentiment analysis (ACSA) has become an essential technology in real-world opinion mining applications, powering platforms such as e-commerce review analytics, customer experience monitoring, and market intelligence systems \cite{DL-absa}. It aims to determine the sentiment polarity towards predefined aspect categories within a given text. This category-specific sentiment understanding enables enterprises to pinpoint actionable insights, detect domain-specific user concerns, and support wiser decision-making. Despite its practical importance, most existing methods primarily focus on learning coarse-grained sentiment polarity (i.e., positive, neutral, negative), while overlooking capturing the richer affective signals expressed by users, such as anger, disgust, joy, or surprise \cite{ABEA-multi}. These affective expressions provide deeper insight into the sentiment reasoning process.

Recent research has begun to explore finer affective dimensions beyond sentiment polarity. Early studies such as \cite{ABEA,ABEA-multi,EmoGRACE} introduced aspect-based emotion analysis (ABEA), advancing the task from simple positive/negative/neutral predictions to recognizing emotions associated with aspect terms or aspect categories. However, these studies typically formulate ABEA as a set of independent tasks, training separate classifiers for aspect term extraction (ATE), aspect category classification (ACC), or aspect emotion classification (AEC). These models fail to jointly learn the interaction between sentiment polarity and emotion signals, hindering the model’s capacity to understand and generalize affective expressions beyond coarse-grained polarity.

A prominent technique to address task separation is multi-task learning (MTL), which leverages shared representations across multiple related tasks to improve generalization performance compared to training models for each task independently \cite{MTL,MTL-lai}. Prior studies, such as \cite{MTL-multi,MTL-SET} demonstrated that utilizing MTL architectures for jointly handling sentiment analysis and emotion analysis can mutually enhance the model's performance. However, these approaches rely solely on existing emotion annotations, without mechanisms for verifying or refining the quality of the emotion labels. Low-quality or noisy emotion labels can limit the model's ability to perform effective emotion-aware sentiment reasoning.

The emergence of large language models (LLMs) has significantly changed the patterns of text understanding and generation. LLMs have been pre-trained on diverse text corpora, making them competent for emotion generation and affective reasoning even when sentiment expressions are implicitly expressed \cite{da}. This makes them ideal for generating category-emotion pairs alongside category-sentiment pairs within a unified generative framework. However, existing ACSA and ABEA research has rarely leveraged the affective richness of LLMs to jointly model sentiment and emotion at the aspect-category level. Moreover, even though LLMs are capable of producing emotional expressions, their outputs can still contain inconsistencies or hallucinations without further refinement.

With the above motivations, we introduce an emotion-enhanced multi-task learning framework that integrates category-level emotions into ACSA. Specifically, we leverage an LLM to generate emotion labels grounded in Ekman's emotion theory for each aspect category within a sentence. The model learns both category-sentiment pairs and category-emotion pairs from the same input sentence. This design allows our ACSA model to capture nuanced affective signals that traditional polarity-only models overlook. To ensure the quality and psychological validity of the generated emotions, we further incorporate a Valence-Arousal-Dominance (VAD) based refinement strategy. We compare the LLM-generated emotions with canonical VAD-mapped emotion labels. If the LLM-generated emotion is inconsistent with the VAD-mapped emotion, we re-annotate the emotion using an LLM-driven refinement procedure. This provides high-quality emotion supervision for the multi-task learning process.

In summary, our contributions are as follows:

\begin{itemize}
  \item We introduce an LLM-based multi-task learning approach that jointly learns sentiment polarity and category-specific emotions based on Ekman’s emotion theory, enabling the model to learn richer affective representations for ACSA.
  \item Our method incorporates a Valence-Arousal-Dominance based refinement strategy that enhances the consistency and quality of generated emotions through affective-space mapping and LLM-driven re-annotation.
  \item The experimental results on benchmark datasets demonstrate that our approach achieves significant performance gains in ACSA, validating the effectiveness of integrating emotional reasoning into ACSA.
\end{itemize}

\section{Related Work}

\label{sec-related_work}

\subsection{Aspect Category Sentiment Analysis}

Aspect category sentiment analysis (ACSA) focuses on identifying sentiment polarities associated with predefined aspect categories, enabling systems to understand user opinions at a finer granularity. Early studies treat ACSA as a classification-based approach, which typically decomposes the task into two subtasks: aspect category detection (ACD) and aspect-category sentiment classification (ACSC), and unifies them under a shared discriminative framework. For instance, AddOneDim-BERT \cite{AddOneDim-BERT} unifies category occurrence and sentiment prediction by adding one dimension to the sentiment label space to predict the presence or absence of each category. This converts the task into a multi-class classification problem for each category. AC-MIMLLN \cite{AC-MIMLLN} generates category-specific representations, which are the weighted sum of word representations, based on weights provided by an auxiliary ACD task, and is used for the sentiment classification prediction. ECAN \cite{ECAN} applies a parallel attention mechanism to the coherence-aware representation to extract distinct category disentanglement features and sentiment disentanglement features, addressing the problem of entanglement of multiple aspect categories and sentiments within a single sentence.

To leverage explicit graph structures for modeling relations and dependencies, some studies employ graph-based approaches to represent internal relationships. For instance, Hier-GCN-BERT \cite{Hier-BERT} aims to capture the inner-relations among multiple categories (e.g., co-occurrence frequency) using a lower-level graph convolutional network (GCN), and the inter-relations between aspect categories and category-oriented sentiments using a higher-level GCN. AAGCN \cite{AAGCN} regards each aspect as a pivot to derive ``aspect-aware words" highly related to it. It employs a Beta Distribution to determine the aspect-aware weight for each of these words. These aspect-aware words and their derived weights are then used to construct the graph for learning contextual sentiment dependencies.

With the rise of large language models (LLMs), recent work has framed ACSA as a generative or instruction-following task. For instance, ACSA-gen \cite{ACSA-gen} casts the ACSA task into a natural language generation task using a sequence-to-sequence framework. The encoder takes the input sentence, and the decoder generates a natural language sentence corresponding to the output. LEGO-ABSA \cite{LEGO-ABSA} unifies multiple ABSA tasks using task-assemblable prompts built upon the T5 backbone to facilitate multi-task training and task transfer. MARG \cite{MARG} addresses the challenge of annotation differences in ACSA by utilizing a rule-based prompt generation system, an AutoRegressive PLM, and a specialized majority rules module that incorporates local ensemble and global refinement via label prior knowledge. However, existing ACSA research focuses primarily on predicting sentiment polarities such as positive, negative, or neutral, ignoring the richer emotional dimensions that often underlie user opinions.

\begin{figure*}[htbp]
  \centering
 \includegraphics[width=0.9\textwidth]{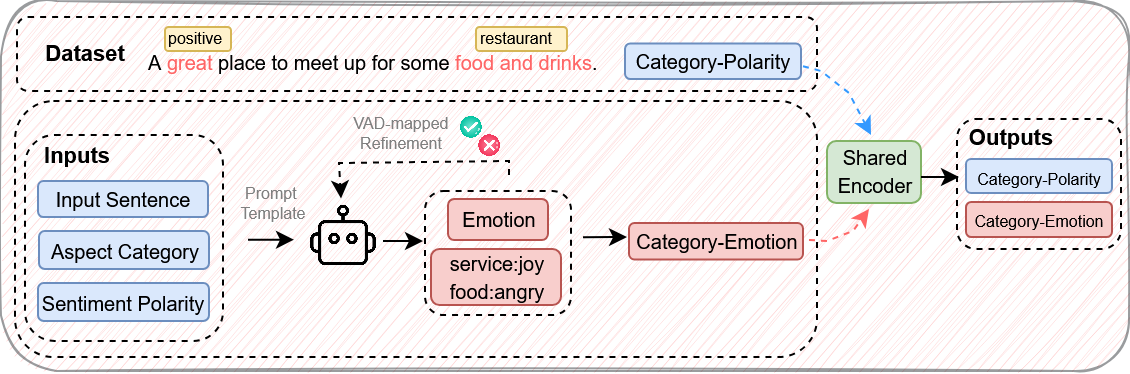}

  \caption{Overview of our emotion-enhanced multi-task ACSA framework.}

  \label{pic:overview}

\end{figure*}

\subsection{Emotion Modeling}

Emotion modeling in natural language processing has been dominated by two influential paradigms. The first is the discrete emotion category framework represented by Ekman’s six basic emotions \cite{Ekman} (anger, disgust, fear, joy, sadness, and surprise), which assumes that human affect can be decomposed into a set of universal, grounded categories. The second paradigm adopts a dimensional perspective, rooted in the psychological model of \cite{VAD}, which conceptualizes affective states along continuous axes of valence, arousal, and dominance (VAD). Remarkably, research \cite{mapped} further demonstrated that emotional states can be comprehensively defined by three independent and bipolar dimensions: pleasure-displeasure, degree of arousal, and dominance-submissiveness, providing numerical valence-arousal-dominance coordinates for Ekman’s basic emotions.

Within aspect-based sentiment analysis (ABSA), more recent research has shifted from coarse polarity to richer affective understanding that integrates emotion signals at the aspect level. Prior works, such as EmoGRACE \cite{EmoGRACE} integrate emotion by adapting the ABSA model GRACE to perform the joint tasks of aspect term extraction (ATE) and aspect emotion classification (AEC), where AEC involves assigning categorical emotion labels (such as happiness, anger, sadness, and fear) to the extracted target words. \cite{ABEA-LSTM} integrates emotion by utilizing a hybrid model to mine text from tweets about the Ukraine-Russia conflict, classifying and displaying the results not only in terms of sentiment but also into specific emotion categories such as anger, optimism, joy, and sadness, providing a more detailed analysis than broad sentiment alone. \cite{ABEA-multi} focuses on annotating and classifying customer comments according to fine-grained emotional categories and the emotional dimensions of valence and arousal at the aspect level. However, these studies typically decompose the task into separate subtasks, such as ATE and AEC. They train independent classifiers for each component and do not jointly learn category-sentiment and category-emotion relations, preventing them from leveraging the complementary interaction between polarity and affective states.

\subsection{Multi-task Learning}

Multi-task learning (MTL) is a paradigm in which a model is trained to solve multiple related tasks simultaneously, allowing shared representations to improve generalization \cite{MTL}. In the context of sentiment analysis, early studies have explored integrating sentiment and emotion prediction through multi-task learning, demonstrating the potential benefits of joint modeling. For instance, \cite{MTL-SET} utilizes an MTL setup by training a shared BERT-based encoder with multiple classification heads to jointly model the primary task of Hate Speech and Offensive Language (HOF) detection alongside three related auxiliary tasks: sentiment classification, emotion classification, and target classification to exploit their correlations and achieve improved generalization for the HOF class. \cite{MTL-multi} proposes a deep MTL framework that simultaneously learns and classifies the two related affective tasks of multi-modal sentiment analysis and emotion recognition by creating a shared representation from a contextual inter-modal (CIM) attention mechanism applied across text, acoustic, and visual inputs, thus leveraging the tasks' inter-dependence to enhance predictive performance.

However, existing MTL approaches that combine emotion analysis face a challenge in terms of the quality of emotion annotation. As \cite{MTL-SET} acknowledges that the primary emotion corpus used in their task is noisy, containing inconsistent emotion labels. Prior work relies on externally provided emotion annotations, which lack mechanisms to refine or validate the correctness of emotion labels. As a result, these MTL models may not be able to fully capture the nuanced interaction between polarity expressions and affective signals. In contrast, to guarantee the reliability of the emotion label, we introduce a refinement strategy based on Valence-Arousal-Dominance (VAD) dimensional mappings. This refinement mechanism filters out inconsistent emotion labels, producing high-quality auxiliary supervision for our multi-task model.

\section{Method}
We present the emotion-enhanced multi-task ACSA framework in this section. The overview of our methodology is shown in Figure \ref{pic:overview}.

\subsection{Task Definition}

Aspect category sentiment analysis (ACSA) aims to determine the sentiment polarity associated with predefined aspect categories within a sentence. Given an input sentence $S$ and a set of aspect categories $\mathcal{C} = \{c_1, c_2, \dots, c_m\}$, the goal of ACSA is to predict all category-related sentiment polarities $\{(c_i, y_i)\}_{i=1}^m$ that appear in $S$, where $y_i \in \{\text{positive}, \text{neutral}, \text{negative}\}$.

\subsection{Category-level Emotion Generation}

Conventional ACSA methods primarily focus on modeling the sentiment polarity, overlooking the rich emotional signals that underlie human affective expression. According to Ekman's emotion theory \cite{Ekman}, human emotions can be categorized into six universal types: anger, disgust, fear, joy, sadness, and surprise, which provide a more fine-grained affective representation than coarse-grained sentiment polarities. 

\begin{figure}[htbp]

  \centering

  \includegraphics[width=0.53\textwidth]{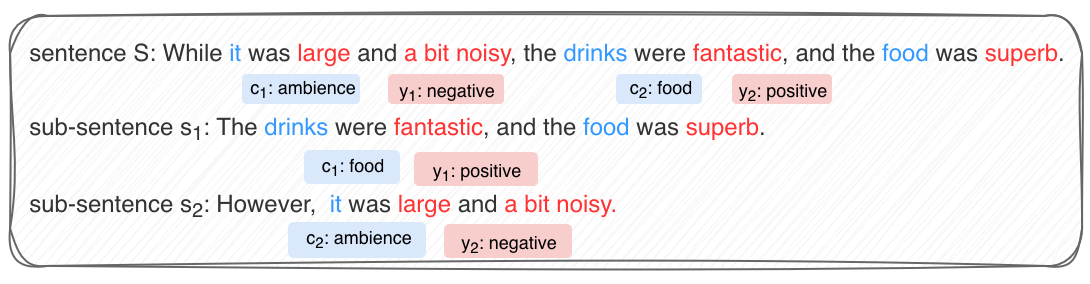}

  \caption{An example of using an LLM to decompose sentences.}

  \label{pic:sub}

\end{figure}

To integrate emotional knowledge into ACSA, we employ large language models (LLMs) to generate category-oriented emotion annotations. However, each sentence may contain multiple category-sentiment pairs, leading to overlapping semantic expressions, which may result in LLMs generating ambiguous emotions that reflect the overall emotion of the sentence rather than the affective state corresponding to each category. To resolve this issue, we first utilize the LLM to decompose each sentence $S$ into multiple sub-sentences $\{s_i\}_{i=1}^m$, where each $s_i$ contains linguistic context that reflects only the corresponding $(c_i, y_i)$ pair. This decomposition ensures that the LLM-generated emotion for each sub-sentence $s_i$ accurately captures the affective expression associated with a single aspect category. Figure \ref{pic:sub} shows an example of using an LLM to decompose a sentence.

\begin{figure}[htbp]

  \centering

  \includegraphics[width=0.5\textwidth]{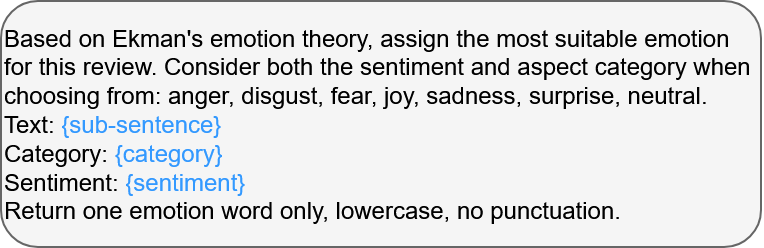}

  \caption{Prompt template employed for the emotion generation.}

  \label{pic:emotion}

\end{figure}

After obtaining the category-specific sub-sentences $\{s_i\}_{i=1}^{m}$, we prompt the LLM to infer the most appropriate emotion $e_i$ for each $(s_i, c_i, y_i)$ triplet from an extended Ekman's six basic emotions set:

\[
\mathcal{E} = \{\text{anger}, \text{disgust}, \text{fear}, \text{joy}, \text{sadness}, \text{surprise}, \text{neutral}\}.
\]

The inclusion of neutral is to align with the original sentiment polarity. The structured prompt template for emotion generation is shown in Figure \ref{pic:emotion}.

\subsection{VAD-based Emotion Refinement}

\label{vad-refine}


To further ensure the reliability of the generated emotions, we revise the emotion annotations using the Valence-Arousal-Dominance (VAD) dimensional emotion theory \cite{VAD}. Specifically, we employ the EmoBank corpus \cite{EmoBank}, which provides human-annotated VAD scores, to fine-tune a DeBERTa model \cite{deberta} to predict continuous VAD values for each sub-sentence $s_i$, $\hat{v}_i, \hat{a}_i, \hat{d}_i = \mathrm{DeBERTa}(s_i)$, where $\hat{v}_i, \hat{a}_i, \hat{d}_i$ correspond to valence, arousal, and dominance ratings on a five-point scale.

\begin{table}[!ht]
\caption{The valence, arousal, and dominance values of emotions adopted from \cite{mapped,centroid}.}
   \label{tab:vad}
    \centering   
\begin{tabular}{llll}
\hline
                              & \multicolumn{1}{c}{Valence} & \multicolumn{1}{c}{Arousal} & \multicolumn{1}{c}{Dominance} \\
                              \hline
\multicolumn{1}{c}{Anger}     & \multicolumn{1}{c}{$-0.43$ }   & \multicolumn{1}{c}{0.67}    & \multicolumn{1}{c}{0.34}      \\
\multicolumn{1}{c}{Disgust}   & \multicolumn{1}{c}{$-0.60$}   & \multicolumn{1}{c}{0.35}    & \multicolumn{1}{c}{0.11}      \\
\multicolumn{1}{c}{Fear}      & \multicolumn{1}{c}{$-0.64$}   & \multicolumn{1}{c}{0.6}     & \multicolumn{1}{c}{$-0.43$}     \\
\multicolumn{1}{c}{Happiness} & \multicolumn{1}{c}{0.76}    & \multicolumn{1}{c}{0.48}    & \multicolumn{1}{c}{0.35}      \\
\multicolumn{1}{c}{Surprise}  & \multicolumn{1}{c}{0.4}     & \multicolumn{1}{c}{0.67}    & \multicolumn{1}{c}{$-0.13$}     \\
\multicolumn{1}{c}{Sadness}   & \multicolumn{1}{c}{$-0.63$}   & \multicolumn{1}{c}{0.27}    & \multicolumn{1}{c}{$-0.33$}     \\
\hline
\end{tabular}
\end{table}

To normalize these scores for mapping into categorical emotions, we convert each dimension into the range $[-1,1]$ using 

$v = \frac{\hat{v}_i - 3}{2}, \qquad
a = \frac{\hat{a}_i - 3}{2}, \qquad
d = \frac{\hat{d}_i - 3}{2}$ to obtain the final VAD values $(v, a, d)$. We map the continuous VAD predictions to discrete emotion categories using a nearest-centroid classifier in VAD space. Each Ekman emotion is associated with a centroid $(v^*, a^*, d^*)$ defined by Russell and Mehrabian \cite{mapped}, as shown in Table \ref{tab:vad}. The emotion mapped from VAD value $e^{\mathrm{vad}}_i$ is determined as the label with the minimum Euclidean distance:

\[
e^{\mathrm{vad}}_i
= \arg\min_{e \in \mathcal{E}}
\left\| (v, a, d) - (v^*_{e}, a^*_{e}, d^*_{e}) \right\|_2^2.
\]

\begin{figure}[htbp]

  \centering

  \includegraphics[width=0.5\textwidth]{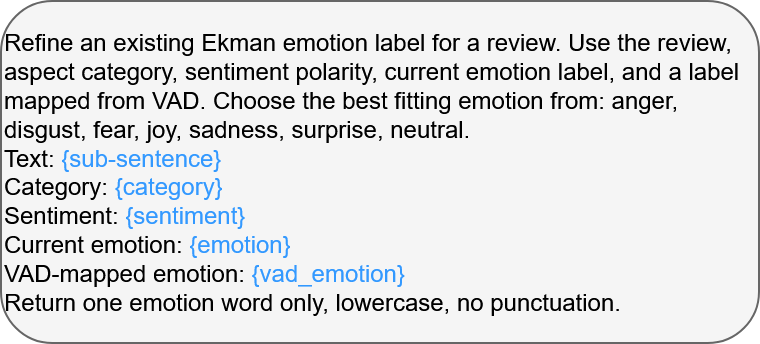}

  \caption{Prompt template employed for the revised emotion generation.}

  \label{pic:refine}

\end{figure}


To obtain high-quality emotion labels, we retain the LLM-generated emotion if $e_i = e^{\mathrm{vad}}_i$, as it reflects both contextual semantics and affective grounding. However, if the LLM-generated emotion is inconsistent with the emotion mapped from VAD value $e^{\mathrm{vad}}_i$, we perform a refinement process by prompting the LLM with specific information, including the sub-sentence $s_i$, its category $c_i$, its sentiment $y_i$, the original LLM-generated emotion $e_i$, and the VAD-mapped emotion $e^{\mathrm{vad}}_i$ to guarantee the affective consistency of the generated emotion labels. The prompt template for revised emotion generation is shown in Figure \ref{pic:refine}.

\subsection{Multi-task Learning}

To enable the model's capacity to understand fine-grained affective semantics grounded in Ekman’s emotions, we adopt a multi-task learning (MTL) framework where an auxiliary emotion analysis task is jointly optimized with the primary sentiment analysis task. The auxiliary task enhances the model's ability to understand nuanced emotional information that traditional ACSA tasks fail to capture, thereby improving its generalization capacity and affective reasoning.

Given an input sentence $x$, the objective of model training is to generate two types of outputs. The first type is a set of category-sentiment pairs, a sequence that describes all aspect categories present in the sentence and their corresponding sentiment polarities. For example: \textit{``LAPTOP\#QUALITY:negative"; ``LAPTOP\#GENERAL:positive"}. The second type is a set of category-emotion pairs, a sequence that describes the emotion label associated with each category, using the revised emotion annotations generated by the VAD-revised pipeline in Section \ref{vad-refine}. For example: \textit{``LAPTOP\#QUALITY:disgust"; ``LAPTOP\#GENERAL:joy"}.

We employ a sequence-to-sequence model \(f_\theta\), which shares parameters for both tasks but differs only in the target sequence. For an input \(x\), the model generates conditional token distributions over the vocabulary via $p_\theta\!\bigl(y_t \mid y_{<t},\, t(\,x\,)\bigr) $.

For the category-sentiment task, we optimize the standard negative log-likelihood for each sentence. The loss for category-sentiment prediction is expressed as follows:

\[
\mathcal{L}_{\mathrm{sen}}(\theta)
\;=\;
-\mathbb{E}
\sum_{t=1}^{T}
\log p_\theta\!\bigl(y^{\mathrm{sen}}_t \,\big|\, y^{\mathrm{sen}}_{<t},\, t^{\mathrm{sen}}(x)\bigr)
\]

Similarly, for the category-emotion task with revised emotions that are grounded in dimensional emotion theory, the training loss is expressed as the following equation:

\[
\mathcal{L}_{\mathrm{emo}}(\theta)
\;=\;
-\mathbb{E}
\sum_{t=1}^{T}
\log p_\theta\!\bigl(y^{\mathrm{emo}}_t \,\big|\, y^{\mathrm{emo}}_{<t},\, t^{\mathrm{emo}}(x)\bigr)
\]

where \(y^{\mathrm{sen}},y^{\mathrm{emo}}\) represent serialized targets and \(t^{\mathrm{sen}}(\cdot),t^{\mathrm{emo}}(\cdot)\) refer to input sequences.

The final training objective is a weighted combination of the two losses. The overall joint optimization objective is presented as follows:

\[
\mathcal{L}(\theta)
\;=\;
\alpha\,\mathcal{L}_{\mathrm{sen}}(\theta)
\;+\;
\bigl(1-\alpha\bigr)\,\mathcal{L}_{\mathrm{emo}}(\theta)
\]

where $\alpha \in [0,1]$ controls the balance between sentiment learning and emotion learning. A higher value of $\alpha$ places greater emphasis on traditional ACSA performance, while smaller values encourage the model to rely more heavily on emotion modeling. In our experiments, we fine-tune $\alpha$ based on the validation performance to ensure that both tasks contribute meaningfully to the shared parameter updates.

Jointly optimizing \(\mathcal{L}_{\mathrm{sen}}\) and \(\mathcal{L}_{\mathrm{emo}}\) enables the shared encoder-decoder representation to capture affective semantics that cannot be captured by coarse polarity alone. The auxiliary supervision on emotional labels encourages the model to differentiate, for example, \textit{joy} vs.\ \textit{surprise} within the same positive polarity, enhancing the model's capacity to understand affective semantics beyond coarse-grained polarity.

\section{Experiment}

\subsection{Datasets and Evaluation Metrics}

 We use the Restaurant and Laptop datasets from SemEval-2015 Task 12 \cite{sem15} and SemEval-2016 Task 5 \cite{sem16} to conduct our experiments. We compare our method with two groups of baselines: classical ACSA baselines and instruction-based fine-tuning methods. All results are reported as Precision (P), Recall (R), and F1 score (F1).

\subsection{Baselines}

In the ACSA task, we compare our approach with the following methods.

\begin{itemize}
  \item \textbf{ACSA Baselines:}
  \begin{itemize}
    \item[-] \textbf{BERT}: This method establishes separate pipelines for aspect category detection and sentiment polarity classification, utilizing BERT \cite{BERT} as the review encoder for both individual subtasks.
    \item[-] \textbf{Cartesian-BERT}: This method utilizes BERT as the sentence encoder and employs the Cartesian product method.
    \item[-] \textbf{AddOnedim-BERT} \cite{AddOneDim-BERT}: This method unifies category presence or absence and sentiment prediction into a single multi-class classification problem by extending the sentiment label space to include N/A.
    \item[-] \textbf{AS-DATJM} \cite{AS-DATJM}: This method employs an undirected dependency graph to construct syntactic dependency relations and leverages a dual-attention mechanism to establish the interrelation between aspect category and aspect sentiment.
    \item[-] \textbf{Hier-Trans-BERT} \cite{Hier-BERT}: This method utilizes a Transformer module to establish the inner-relation between aspect categories and the inter-relation between categories and sentiment polarity classification.
    \item[-] \textbf{PBJM} \cite{PBJM}: This method decomposes the ACSA task into category prediction using prompt learning with a verbalizer and sentiment polarity prediction as binary classification.
  \end{itemize}
\end{itemize}

\begin{itemize}
  \item \textbf{Instruction-based Fine-tuning:}
  \begin{itemize}
    \item[-] \textbf{Flan-T5}: This method reformats the input into an instruction and generates the corresponding category-sentiment label as the output sequence.
    \item[-] \textbf{Tk-Instruct-def} \cite{Tk-INSTRUCT}: A T5-based model that is specifically fine-tuned to follow natural language definitions (def) and instructions for NLP tasks. Similar to Flan-T5, it reformulates ACSA into a unified generative task using pre-defined instructions.
    \item[-] \textbf{Tk-Instruct-def-pos} \cite{Tk-INSTRUCT}: A variant of the Tk-Instruct framework, typically combines the task definition (def) and positive demonstration examples (pos).
  \end{itemize}
\end{itemize}

\subsection{Models and Hyperparameters}

\label{sec:exp-models}

For the auxiliary task, we use GPT-4o-mini \footnote{\url{https://platform.openai.com/docs/models/}} to generate revised emotions (anger, disgust, fear, joy, sadness, surprise, neutral) for each sentence that contains a set of category-sentiment pairs. We adopt Flan-T5 (3B) \cite{flan-t5}, a sequence-to-sequence backbone as the multi-task learner. The model jointly learns to predict category-sentiment pairs and category-emotion pairs per input sentence. Other instruction-based fine-tuning models, such as Flan-T5, Tk-Instruct-def, and Tk-Instruct-def-pos, are trained with the same model size (3B) to ensure a faithful comparison across methods.

We train all models using AdamW, a learning rate of 3e-5, a batch size of 4, and 10 epochs. We report the average experimental result over three randomly selected seeds. We reserve 10\% of the training data as a validation split. $\alpha$ is determined via a grid search over the set of $\{0.1,0.2,0.3,0.4,0.5,0.6,0.7,0.8,0.9\}$ and our reported multi-task results use $\alpha=0.6$ as when $\alpha=0.6$ consistently yields the best validation performance, offering the best trade-off between leveraging coarse-grained polarity and fine-grained affective emotions.

\subsection{Main Results}

\begin{table*}[!ht]
\caption{Main results of our model compared with representative baselines on the benchmark dataset. The highest scores are highlighted in bold. $^{\ast}$ refers that the experimental results are based on PBJM \cite{PBJM}.}
   \label{tab:main-results}
    \centering    
\begin{tabular}{lllllllllllll}
\hline
\multicolumn{1}{c}{\multirow{2}{*}{Model}} & \multicolumn{3}{c}{Rest15}                                                                                   & \multicolumn{3}{c}{Rest16}                                                                                   & \multicolumn{3}{c}{Lap15}                                                                                    & \multicolumn{3}{c}{Lap16}                                                                                    \\

\multicolumn{1}{c}{}                       & \multicolumn{1}{c}{P}              & \multicolumn{1}{c}{R}              & \multicolumn{1}{c}{F1}             & \multicolumn{1}{c}{P}              & \multicolumn{1}{c}{R}              & \multicolumn{1}{c}{F1}             & \multicolumn{1}{c}{P}              & \multicolumn{1}{c}{R}              & \multicolumn{1}{c}{F1}             & \multicolumn{1}{c}{P}              & \multicolumn{1}{c}{R}              & \multicolumn{1}{c}{F1}             \\
\hline
\textbf{ACSA Baselines}                    & \multicolumn{1}{c}{}               & \multicolumn{1}{c}{}               & \multicolumn{1}{c}{}               & \multicolumn{1}{c}{}               & \multicolumn{1}{c}{}               & \multicolumn{1}{c}{}               & \multicolumn{1}{c}{}               & \multicolumn{1}{c}{}               & \multicolumn{1}{c}{}               & \multicolumn{1}{c}{}               & \multicolumn{1}{c}{}               & \multicolumn{1}{c}{}               \\
Pipeline-BERT $^{\ast}$                             & \multicolumn{1}{c}{38.12}          & \multicolumn{1}{c}{70}             & \multicolumn{1}{c}{49.35}          & \multicolumn{1}{c}{43.62}          & \multicolumn{1}{c}{79.06}          & \multicolumn{1}{c}{56.21}          & \multicolumn{1}{c}{36.91}          & \multicolumn{1}{c}{51.62}          & \multicolumn{1}{c}{43.02}          & \multicolumn{1}{c}{31.92}          & \multicolumn{1}{c}{51.56}          & \multicolumn{1}{c}{39.42}          \\
Cartesian-BERT $^{\ast}$ & \multicolumn{1}{c}{72.02} & \multicolumn{1}{c}{49.15} & \multicolumn{1}{c}{58.42} & \multicolumn{1}{c}{74.96} & \multicolumn{1}{c}{63.84} & \multicolumn{1}{c}{68.94} & \multicolumn{1}{c}{73.06} & \multicolumn{1}{c}{21.18} & \multicolumn{1}{c}{32.83} & \multicolumn{1}{c}{64.99} & \multicolumn{1}{c}{27.4}  & \multicolumn{1}{c}{39.54} \\
Addonedim-BERT $^{\ast}$ & \multicolumn{1}{c}{68.84} & \multicolumn{1}{c}{55.86} & \multicolumn{1}{c}{61.67} & \multicolumn{1}{c}{71.75} & \multicolumn{1}{c}{67.95} & \multicolumn{1}{c}{69.79} & \multicolumn{1}{c}{64.13} & \multicolumn{1}{c}{39.57} & \multicolumn{1}{c}{48.94} & \multicolumn{1}{c}{58.83} & \multicolumn{1}{c}{39.49} & \multicolumn{1}{c}{47.23} \\
AS-DATJM $^{\ast}$ & \multicolumn{1}{c}{66.35} & \multicolumn{1}{c}{50.52} & \multicolumn{1}{c}{57.21} & \multicolumn{1}{c}{70.88} & \multicolumn{1}{c}{60.35} & \multicolumn{1}{c}{65.19} & \multicolumn{1}{c}{58.91} & \multicolumn{1}{c}{40.28} & \multicolumn{1}{c}{47.76} & \multicolumn{1}{c}{57.29} & \multicolumn{1}{c}{36.7} & \multicolumn{1}{c}{44.71}         \\
Hier-Trans-BERT $^{\ast}$                            & \multicolumn{1}{c}{70.22}          & \multicolumn{1}{c}{59.96}          & \multicolumn{1}{c}{64.67}          & \multicolumn{1}{c}{73.25}          & \multicolumn{1}{c}{73.21}          & \multicolumn{1}{c}{73.45}          & \multicolumn{1}{c}{65.63}          & \multicolumn{1}{c}{51.95}          & \multicolumn{1}{c}{57.79}          & \multicolumn{1}{c}{58.06}          & \multicolumn{1}{c}{48.29}          & \multicolumn{1}{c}{52.52}          \\
PBJM $^{\ast}$                                  & \multicolumn{1}{c}{75.07}          & \multicolumn{1}{c}{61.48}          & \multicolumn{1}{c}{67.58}          & \multicolumn{1}{c}{76.53}          & \multicolumn{1}{c}{73.6}           & \multicolumn{1}{c}{75.03}          & \multicolumn{1}{c}{72.23}          & \multicolumn{1}{c}{54.96}          & \multicolumn{1}{c}{62.41}          & \multicolumn{1}{c}{62.63}          & \multicolumn{1}{c}{50.08}          & \multicolumn{1}{c}{55.62}          \\
\hline
\textbf{Instruction-based Fine-tuning}     & \multicolumn{1}{c}{}               & \multicolumn{1}{c}{}               & \multicolumn{1}{c}{}               & \multicolumn{1}{c}{}               & \multicolumn{1}{c}{}               & \multicolumn{1}{c}{}               & \multicolumn{1}{c}{}               & \multicolumn{1}{c}{}               & \multicolumn{1}{c}{}               & \multicolumn{1}{c}{}               & \multicolumn{1}{c}{}               & \multicolumn{1}{c}{}               \\
Flan-T5                                    & \multicolumn{1}{c}{83.01}          & \multicolumn{1}{c}{77.59}          & \multicolumn{1}{c}{80.21}          & \multicolumn{1}{c}{84.46}          & \multicolumn{1}{c}{84.7}           & \multicolumn{1}{c}{84.58}          & \multicolumn{1}{c}{77.83}          & \multicolumn{1}{c}{74.08}          & \multicolumn{1}{c}{75.89}          & \multicolumn{1}{c}{66.22}          & \multicolumn{1}{c}{66.5}           & \multicolumn{1}{c}{66.36}          \\
Tk-Instruct-def                         & \multicolumn{1}{c}{79.92}          & \multicolumn{1}{c}{74.9}           & \multicolumn{1}{c}{77.33}          & \multicolumn{1}{c}{83.79}          & \multicolumn{1}{c}{85.6}           & \multicolumn{1}{c}{84.69}          & \multicolumn{1}{c}{76.45}          & \multicolumn{1}{c}{73.55}          & \multicolumn{1}{c}{74.97}          & \multicolumn{1}{c}{65.42}          & \multicolumn{1}{c}{65.42}          & \multicolumn{1}{c}{65.42}          \\
Tk-Instruct-def-pos                    & \multicolumn{1}{c}{82.85}          & \multicolumn{1}{c}{77.34}          & \multicolumn{1}{c}{80}             & \multicolumn{1}{c}{83.24}          & \multicolumn{1}{c}{82.91}          & \multicolumn{1}{c}{83.07}          & \multicolumn{1}{c}{75.47}          & \multicolumn{1}{c}{72.29}          & \multicolumn{1}{c}{73.84}          & \multicolumn{1}{c}{64.69}          & \multicolumn{1}{c}{64.04}          & \multicolumn{1}{c}{64.37}          \\
\hline
Ours                                       & \multicolumn{1}{c}{\textbf{84.92}} & \multicolumn{1}{c}{\textbf{78.62}} & \multicolumn{1}{c}{\textbf{81.65}} & \multicolumn{1}{c}{\textbf{85.06}} & \multicolumn{1}{c}{\textbf{86.14}} & \multicolumn{1}{c}{\textbf{85.01}} & \multicolumn{1}{c}{\textbf{79.01}} & \multicolumn{1}{c}{\textbf{77.34}} & \multicolumn{1}{c}{\textbf{78.17}} & \multicolumn{1}{c}{\textbf{67.74}} & \multicolumn{1}{c}{\textbf{68.16}} & \multicolumn{1}{c}{\textbf{67.95}} \\
\hline
\end{tabular}
\end{table*}

\subsubsection{Comparison with ACSA baselines}  

Table \ref{tab:main-results} reports the main results on four benchmarks. Our method consistently outperforms all classical ACSA baselines on every dataset and metric. For a representative baseline PBJM, our approach gains the F1 score of +14.07\% on Rest15, +9.98\% on Rest16, +15.76\% on Lap15, and +12.33\% on Lap16. This demonstrates that treating ACSA as a generative instruction-tuned task yields stronger performance than classification-only models, and enhancing polarity prediction with fine-grained emotion signals and VAD-mapped refinement produces improvements in the ACSA task.

\subsubsection{Comparison with instruction-based fine-tuning methods}  

Compared to instruction-based generative baselines, Flan-T5 (3B) and two Tk-Instruct (3B) variants, our model still attains the best performance across all datasets and metrics. These consistent improvements indicate that the additional emotion-enhanced supervision with VAD refinement and joint multi-task training enhances the model's capacity to understand and generalize affective signals beyond instruction tuning alone.

\subsection{Ablation Study}

We conduct an ablation study on all four datasets to confirm the contribution of each component in our emotion-enhanced multi-task learning framework. Table \ref{tab:ablation} reports results for our method and two variants: Ours (w/o revise), which removes the VAD-mapped refinement and uses only the raw DeBERTa-VAD-mapped emotion labels; and Ours (w/o emotion), which eliminates the auxiliary emotion generation task, focusing solely on the single task of category sentiment prediction.

\begin{table*}[!ht]
\caption{Results of the ablation study demonstrating the contribution of each component to the final performance. }
\label{tab:ablation}
\centering
\begin{tabular}{lllllllllllll}
\hline
\multicolumn{1}{c}{\multirow{2}{*}{Model}} & \multicolumn{3}{c}{Rest15}                                                        & \multicolumn{3}{c}{Rest16}                                                        & \multicolumn{3}{c}{Lap15}                                                         & \multicolumn{3}{c}{Lap16}                                                         \\
\multicolumn{1}{c}{}                       & \multicolumn{1}{c}{P}     & \multicolumn{1}{c}{R}     & \multicolumn{1}{c}{F1}    & \multicolumn{1}{c}{P}     & \multicolumn{1}{c}{R}     & \multicolumn{1}{c}{F1}    & \multicolumn{1}{c}{P}     & \multicolumn{1}{c}{R}     & \multicolumn{1}{c}{F1}    & \multicolumn{1}{c}{P}     & \multicolumn{1}{c}{R}     & \multicolumn{1}{c}{F1}    \\
\hline
\multicolumn{1}{c}{Ours}                   & \multicolumn{1}{c}{\textbf{84.92}} & \multicolumn{1}{c}{\textbf{78.62}} & \multicolumn{1}{c}{\textbf{81.65}} & \multicolumn{1}{c}{\textbf{85.06}} & \multicolumn{1}{c}{\textbf{86.14}} & \multicolumn{1}{c}{\textbf{85.01}} & \multicolumn{1}{c}{\textbf{79.01}} & \multicolumn{1}{c}{\textbf{77.34}} & \multicolumn{1}{c}{\textbf{78.17}} & \multicolumn{1}{c}{\textbf{67.74}} & \multicolumn{1}{c}{\textbf{68.16}} & \multicolumn{1}{c}{\textbf{67.95}} \\
\multicolumn{1}{c}{Ours (w/o revise)}      & \multicolumn{1}{c}{82.05} & \multicolumn{1}{c}{75.8}  & \multicolumn{1}{c}{78.79} & \multicolumn{1}{c}{84.25} & \multicolumn{1}{c}{83.98} & \multicolumn{1}{c}{84.11} & \multicolumn{1}{c}{76.37} & \multicolumn{1}{c}{75.03} & \multicolumn{1}{c}{75.69} & \multicolumn{1}{c}{66.01} & \multicolumn{1}{c}{66.46} & \multicolumn{1}{c}{66.22} \\
\multicolumn{1}{c}{Ours (w/o emotion)}     & \multicolumn{1}{c}{83.01} & \multicolumn{1}{c}{77.59} & \multicolumn{1}{c}{80.21} & \multicolumn{1}{c}{84.46} & \multicolumn{1}{c}{84.7}  & \multicolumn{1}{c}{84.58} & \multicolumn{1}{c}{77.83} & \multicolumn{1}{c}{74.08} & \multicolumn{1}{c}{75.89} & \multicolumn{1}{c}{66.22} & \multicolumn{1}{c}{66.5}  & \multicolumn{1}{c}{66.36} \\
\hline
\end{tabular}
\end{table*}

\subsubsection{Effect of Emotion Supervision}  

The comparison between Ours and Ours (w/o emotion) experimental results demonstrates the clear benefit of the auxiliary emotion task. Removing emotion supervision consistently yields lower F1 scores across all datasets, with average drops of 1.44\% on Rest15, 0.43\% on Rest16, 2.28\% on Lap15, and 1.59\% on Lap16. These results demonstrate that emotional signals enrich the model's internal representation of category-specific evaluation content, providing important affective information for coarse-grained polarity prediction.

\subsubsection{Effect of VAD-mapped Emotion Refinement}  

Comparing the results of Ours and Ours (w/o revise), it can be found that there are consistent improvements when using the refinement strategy. Removing the refinement stage, which only utilizes raw emotion labels from VAD mapping, leads to clear drops in F1: 2.86\% on Rest15, 0.90\% on Rest16, 2.48\% on Lap15, and 1.73\% on Lap16. The performance gap demonstrates that VAD-mapped labels alone are insufficient, as they might be noisy and sometimes misaligned with the emotional content of sub-sentences. The LLM refinement step acts as a context-aware correction mechanism, utilizing polarity, aspect category, and emotions mapped with affective coordinates. Revised emotion labels provide more reliable supervision, leading to improvements in all evaluation results.

\section{Conclusion}

In this paper, we introduce emotion-enhanced multi-task learning with large language models for aspect category sentiment analysis, addressing two fundamental limitations in prior work: the absence of affective grounding in existing ACSA models and the lack of high-quality emotional supervision in multi-task learning frameworks. Our framework utilizes large language models to generate Ekman-style emotion labels, which are subsequently revised through a VAD-mapped refinement strategy, producing reliable category-emotion annotations. The multi-task architecture jointly learns category-sentiment and category-emotion predictions, enabling the model to internalize richer affective semantics beyond coarse-grained polarity. Extensive experiments over four SemEval benchmarks demonstrate that our approach consistently surpasses both classical ACSA baselines and strong instruction-based fine-tuning methods, underscoring the importance of both affective supervision and quality control in learning robust representations.

\section*{Acknowledgment}
This work was fully supported by a grant from the Research Grants Council of the Hong Kong Special Administrative Region, China (R1015-23); the Research Impact Fund by the Research Grants Council of Hong Kong (Project No. 130272); and the Faculty Research Grants (SDS24A8 and SDS24A19) and the Direct Grant (DR25E8) of Lingnan University, Hong Kong.



\bibliographystyle{IEEEtran}
\bibliography{main}
\vspace{-20pt} 


\begin{IEEEbiography}[{\includegraphics[width=1in, height=1.25in,clip,keepaspectratio]{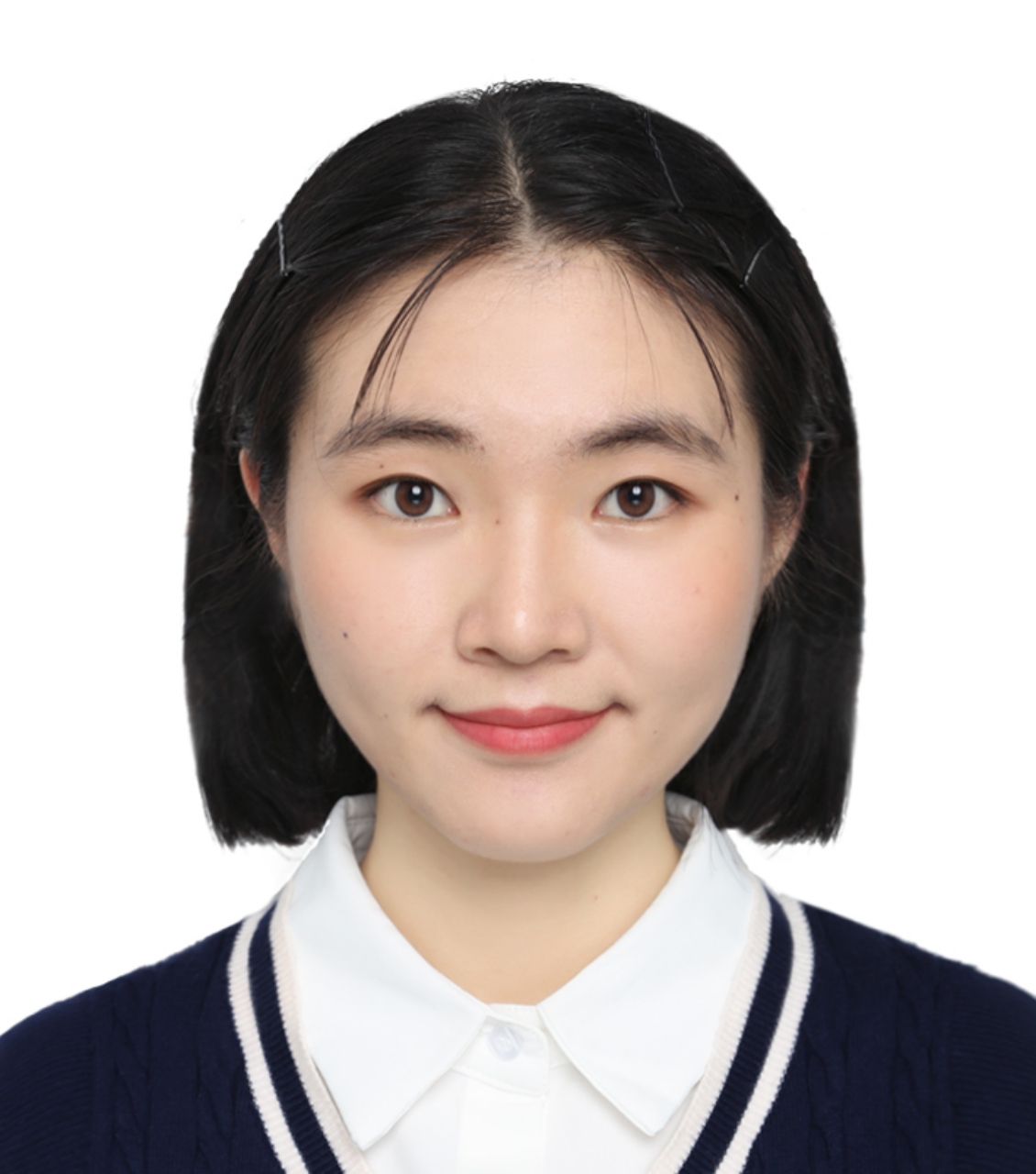}}]{Yaping Chai} is a Ph.D. candidate in the School of Data Science at Lingnan University, Hong Kong, under the supervision of Prof. Joe S. Qin and Prof. Haoran Xie. Her research focuses on large language models, natural language processing, and aspect-based sentiment analysis. Her work explores the development of advanced methods for fine-tuning and evaluating language models in sentiment analysis and related NLP tasks.
\end{IEEEbiography}
\vspace{-25pt}

\begin{IEEEbiography}[{\includegraphics[width=1in,height=1.25in,clip,keepaspectratio]{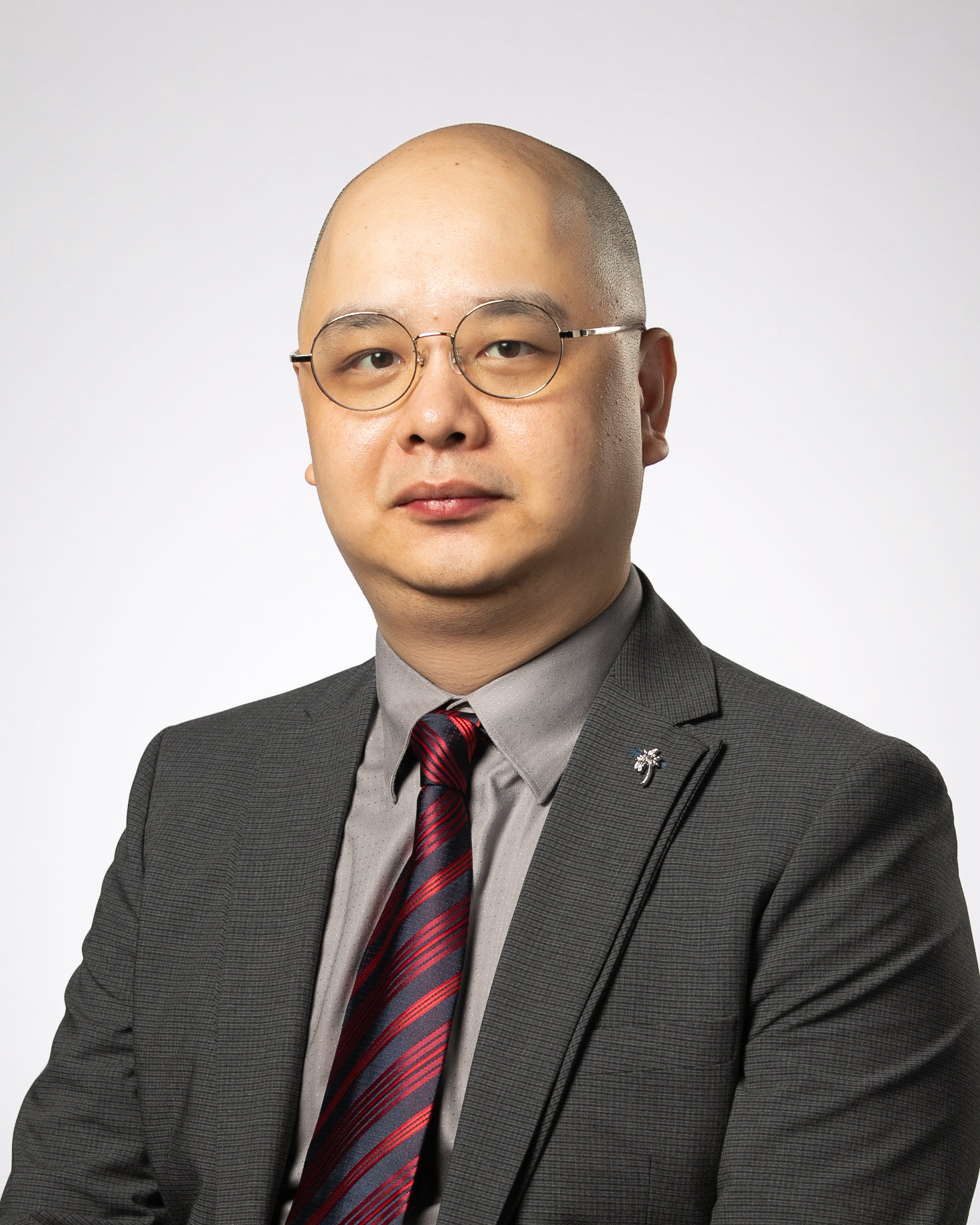}}]{Haoran Xie} (Senior Member, IEEE)
received a Ph.D. degree in Computer Science from City University of Hong Kong and an Ed.D degree in Digital Learning from the University of Bristol. He is currently a Professor and the Person-in-Charge at the Division of Artificial Intelligence, Director of LEO Dr David P. Chan Institute of Data Science, and Associate Dean of the School of Data Science, Lingnan University, Hong Kong. His research interests include natural language processing, large language models, language learning, and AI in education. He has published 450 research publications, including 260 journal articles. He is the Editor-in-Chief of Natural Language Processing Journal, Computers \& Education: Artificial Intelligence, and Computers \& Education: X Reality. He has been selected as the World's Top 2\% Scientists by Stanford University.
\end{IEEEbiography}
\vspace{-25pt}

\begin{IEEEbiography}[{\includegraphics[width=1in,height=1.25in,clip,keepaspectratio]
{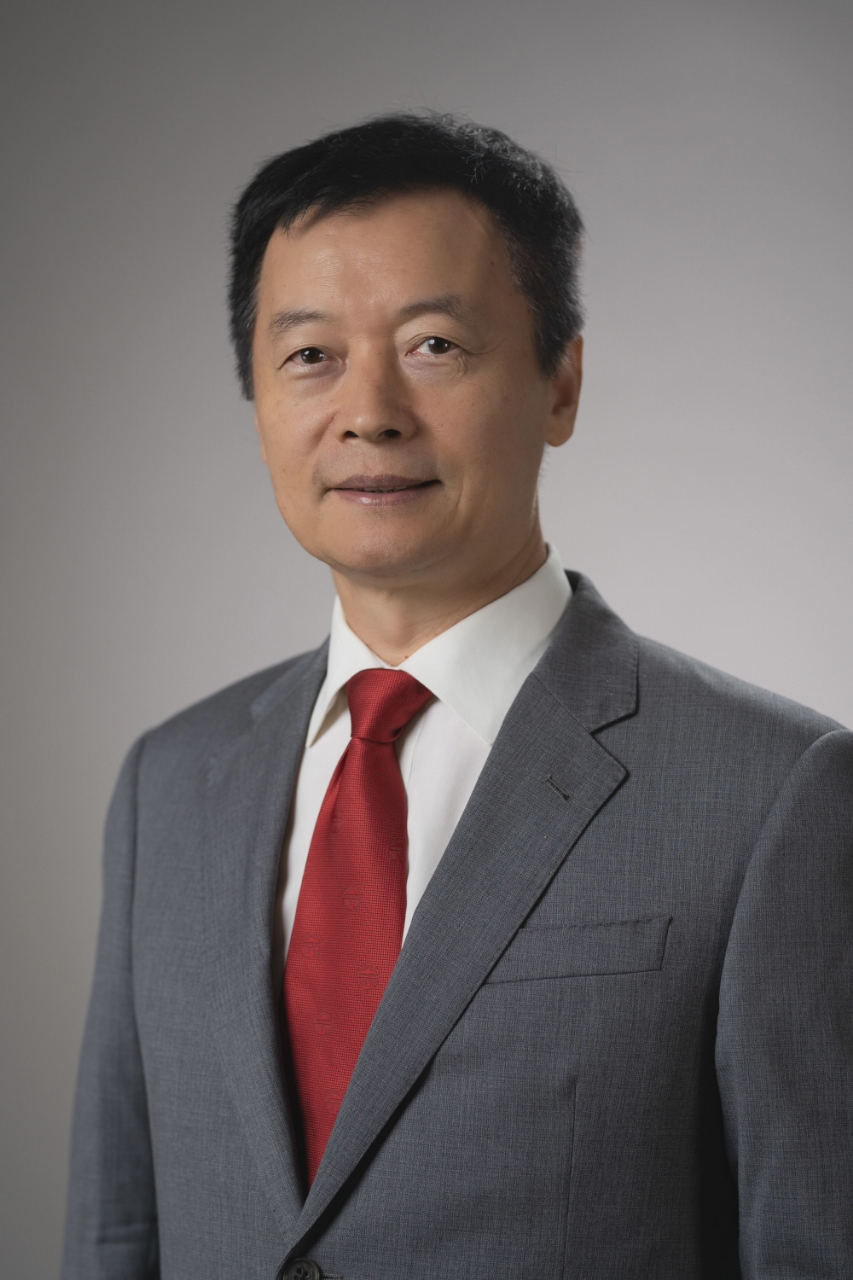}}]
{S. Joe Qin} (Fellow, IEEE) received the B.S. and M.S. degrees in automatic control from Tsinghua University, Beijing, China, in 1984 and 1987, respectively, and the Ph.D. degree in chemical engineering from the University of Maryland, College Park, MD, USA, in 1992. He is currently the Wai Kee Kau Chair Professor and President of Lingnan University, Hong Kong. His research interests include data science and analytics, machine learning, process monitoring, model predictive control, system identification, smart manufacturing, smart cities, and predictive maintenance. Prof. Qin is a Fellow of the U.S. National Academy of Inventors, IFAC, and AIChE. He was the recipient of the 2022 CAST Computing Award by AIChE, 2022 IEEE CSS Transition to Practice Award, U.S. NSF CAREER Award, and NSF-China Outstanding Young Investigator Award. His h-indices for Web of Science, SCOPUS, and Google Scholar are 66, 73, and 89, respectively.
\end{IEEEbiography}
\vfill

\end{document}